\documentclass{article}

\usepackage{microtype}
\usepackage{graphicx}
\usepackage{booktabs} 

\usepackage{hyperref}

\usepackage[accepted]{icml2024}

\usepackage{amsmath}
\usepackage{amssymb}
\usepackage{mathtools}
\usepackage{amsthm}

\usepackage{subcaption}

\usepackage[capitalize,noabbrev]{cleveref}

\usepackage{enumitem}

\theoremstyle{plain}
\newtheorem{theorem}{Theorem}[section]

\theoremstyle{definition}

\theoremstyle{remark}
\newtheorem{remark}[theorem]{Remark}

\usepackage[textsize=tiny]{todonotes}

\usepackage{wrapfig}
\usepackage{multicol}
\icmltitlerunning{Exploiting Estimation Bias in Deep Double Q-Learning}

\begin{document}

\twocolumn[
\icmltitle{Exploiting Estimation Bias in Clipped Double Q-Learning for Continous Control Reinforcement Learning Tasks}



\icmlsetsymbol{equal}{*}

\begin{icmlauthorlist}
\icmlauthor{Niccolò Turcato}{equal,x}
\icmlauthor{Alberto Sinigaglia}{equal,y}
\icmlauthor{Alberto Dalla Libera}{x}
\icmlauthor{Ruggero Carli}{x}
\icmlauthor{Gian Antonio Susto}{x,y}
\end{icmlauthorlist}

\icmlaffiliation{x}{Department of Infomation Engineering, University of Padova, Padova, Italy}
\icmlaffiliation{y}{Human-Inspired Technology Research Center, University of Padova, Padova, Italy}

\icmlcorrespondingauthor{Niccolò Turcato}{niccolo.turcato@phd.unipd.it}

\icmlkeywords{Machine Learning, ICML}

\vskip 0.3in
]



\printAffiliationsAndNotice{\icmlEqualContribution} 

\begin{abstract}
Continuous control Deep Reinforcement Learning (RL) approaches are known to suffer from estimation biases, leading to suboptimal policies.
This paper introduces innovative methods in RL, focusing on addressing and exploiting estimation biases in Actor-Critic methods for continuous control tasks, using Deep Double Q-Learning.
We design a Bias Exploiting (BE) mechanism to dynamically select the most advantageous estimation bias during training of the RL agent. 
Most State-of-the-art Deep RL algorithms can be equipped with the BE mechanism, without hindering performance or computational complexity.
Our extensive experiments across various continuous control tasks demonstrate the effectiveness of our approaches. We show that RL algorithms equipped with this method can match or surpass their counterparts, particularly in environments where estimation biases significantly impact learning. The results underline the importance of bias exploitation in improving policy learning in RL.
\end{abstract}

\section{Introduction}
\label{sec:intro}
Reinforcement Learning (RL) is a tool for training of autonomous agents that can continuously adapt to a changing environment.
Control of continuous action space agents \cite{recht2019continuous_control} is one major area of RL. Popular approaches in the literature include Actor-Critic methods based on temporal difference learning \cite{dann2014policy_temporal_diff}. The current state-of-the-art employs Q-learning \cite{watkins1992qlearning} in which Deep Learning models are used to learn a critic estimate from collected data, while the actor is optimized with policy gradient techniques \cite{lillicrap2015ddpg, fujimoto2018td3, haarnoja2018sac}. 
One critical issue in such methods is the overestimation bias generated by the Q-learning \cite{thrun2014issues_funapprox, fujimoto2018td3}. In \cite{fujimoto2018td3}, the authors introduced Twin Delayed Deep Deterministic Policy Gradient (TD3), a novel algorithm that has the same structure as Deep Deterministic Policy Gradient (DDPG). The algorithm exploits a strategy called Clipped Double Q-Learning (CDQ) to reduce the overestimation of a single critic function. Specifically, two Neural Networks (NNs) are trained independently from one another in order to learn the action value function following the current policy. 
In each learning step, the target for each state-action pair is identical for each NN, and the target is computed with Q-learning using the minimum of the two $Q$ estimates.
CDQ prevents additional overestimation compared to standard Q-learning, at the price of introducing potential underestimation bias. The authors motivate the use of CDQ based on the observation that underestimated actions have much less impact on policy updates than overestimated actions.
Other contributions from \cite{fujimoto2018td3} include (i) the use of target networks for delayed policy updates for both the critic and the actor, which improves learning stability; (ii) regularization of the $Q$ function estimates with Target Policy Smoothing, implementing the intuitive idea that similar actions should be evaluated similarly.
According to \cite{lan2019maxmin_qlearning}, in the context of discrete action spaces, the impact of underestimation may be more problematic than overestimation, depending on the specific environment and task that the agent is dealing with. 
In the case of continuous action spaces, underestimation could potentially lead the agent to adopt overly cautious actions, thus slowing the learning process.
Recent works have tackled the problem of reducing estimation biases by proposing novel strategies, many of which exploit an ensemble of value function estimates. 
The main idea behind these works, which we discuss in \cref{sec:related_work}, consists in trying to reduce the effect of the two estimation biases. Instead, in this work, motivated by numerical results showing that the overestimation bias can lead to higher performance in some circumstances, we discuss how the two biases can be exploited depending on the environment the agent is interacting with.
\\
In this paper, we propose modifications to the TD3 algorithm to leverage estimation biases. Specifically, we introduce a decision layer on top of the algorithm, formulating the choice of bias as a dual-armed bandit problem. This results in the Bias Exploiting - Twin Delayed Deep Deterministic Policy Gradient (BE-TD3) algorithm.
Our results suggest that BE-TD3 can select the most useful estimation bias during the episodes, yielding better policies than TD3, while adding no additional computational burden.
Furthermore, we explore alternative strategies for computing the policy gradient in BE-TD3, diverging from the one originally proposed in \cite{fujimoto2018td3}. 
Our results show that using this richer estimation in the actor updates in some cases can help stabilize learning or allow convergence to higher rewards in environments where TD3 struggled originally.

The paper is structured as follows:
In \cref{sec:related_work}, we discuss recent contributions that are relevant to the scope of this paper, while in \cref{sec:background}, we briefly review the required theoretical background. 
In \cref{sec:critic} we show through an ablation study that the introduction of an overestimation bias in CDQ can improve policy learning, both on a synthetic Markov Decision Process (MDP) and standard continuous control benchmarks. In \cref{sec:bias_exploitation} we discuss the bias selection problem and present BE-TD3, including its results. Finally, \cref{sec:conclusions} concludes the paper by briefly discussing some potential future lines of contributions. \\
To benchmark and compare the algorithms, we use a selection of Mujoco \cite{mujoco} robotics environments from the OpenAI Gym suite \cite{OpenAI_gym}.

\section{Related work}
\label{sec:related_work}
Estimation bias in Value function approximation is an issue that has been addressed in several papers. Q-learning in discrete action space has been shown to suffer from overestimation bias \cite{thrun2014issues_funapprox}. \cite{hasselt2010double_qlearning} presents Double Q-learning as the first possible solution. More recently, Maxmin Q-learning \cite{lan2019maxmin_qlearning} has shown that the use of an ensemble with more than two $Q$ estimates can further reduce this bias and improve Q-learning performances. 
Recent publications regarding control in continuous action spaces address both the underestimation and overestimation bias in Q-learning with the ensembling of $Q$ function estimates \cite{kuznetsov2020controlling_overest, chen2020randomized, wei2022controlling, li2023realistic}.
In \cite{kuznetsov2020controlling_overest}, the authors present Truncated Quantile Critics (TQC) which extends Soft Actor-Critic \cite{haarnoja2018sac} (SAC) using an ensemble of 5 critic estimates to yield a distributional representation of a critic and truncation of the critics' predictions in the critic updates to reduce overestimation bias. TQC also introduces the concept of using the information from multiple Qs in the actor update. Indeed in TQC the actor is updated by optimizing with respect to a sampling of the mean of the ensemble of Qs. 
In \cite{chen2020randomized}, the authors present Randomized Ensembled Double Q-learning (REDQ), which shares a similar structure to TQC, but exploits a larger ensemble with 10 networks and doesn't resort to sampling. In REDQ, the critic estimates are updated multiple times for each step in the environment, 20 times in the presented results.
In \cite{wei2022controlling}, the authors present Quasi-Median Q-learning (QMQ), which uses 4 $Q$ estimates and exploits the quasi-median operator to compute the targets for the critic updates. The authors justify the quasi-median, stating that it is a trade-off between overestimation and underestimation. Instead, the policy gradient is computed with respect to the mean of the Qs.
In \cite{li2023realistic}, the authors propose Realistic Actor-Critic (RAC), an algorithm that targets a balance between value overestimation and underestimation, using an ensemble of 10 $Q$ networks. 
At each step in the environment, the ensemble of $Q$ functions is updated 20 times with targets computed using the mean of the $Q$s minus one standard deviation, while the actor is updated one time by maximizing the mean of the $Q$ functions. In \cite{li2023realistic}, the aforementioned ensemble method is applied on top of both TD3 and SAC, as both methods achieve state-of-the-art performance and sample efficiency comparable to a Model-Based RL approach \cite{janner2019mbpo}.

\remark{The majority of contributions discussed in this section enhance TD3 or SAC by employing large ensembles of $Q$ Networks in the critic or by performing multiple steps of training at each time step, thereby introducing extra computational complexity. In contrast, our work concentrates on enhancing Clipped Double Q-learning by leveraging estimation bias, without imposing any additional computational burden.}

\section{Background}
\label{sec:background}
Reinforcement Learning can be modeled as a Markov Decision Process (MDP), defined by the tuple $(\mathcal{S}, \mathcal{A}, \mathcal{P}, R, \gamma)$. 
$\mathcal{S}$ and $\mathcal{A}$ are the state space and the action space, respectively. Both $\mathcal{S}$ and $\mathcal{A}$ are continuous, thus the transition density function, $\mathcal{P}$ is continuous as well and formalized as $\mathcal{P}: \mathcal{S} \times \mathcal{A} \times \mathcal{S} \rightarrow [0, \infty)$. $r: \mathcal{S} \times \mathcal{A} \times \mathcal{S} \rightarrow \mathbb{R}$ is the random variable describing the reward function, and $\gamma \in [0,1]$ is the discount factor.
A policy $\mu$ at each time-step $t$ is a function that maps from the current state $s_t \in \mathcal{S}$ to an action $a_t \in \mathcal{A}$ with respect to the conditional distribution $\mu(a_t|s_t)$. In our case, such policy is parameterized by parameters $\phi$, thus $\mu_\phi(a_t|s_t)$ will be a Neural Network.
\newline
Reinforcement Learning is concerned with finding the policy that maximizes the expected discounted sum of rewards $R_0 = \mathbb{E_{\mu_\phi}}\left[\sum_{t=0}^\infty \gamma^t r_t\right]$ of a specific MDP. To do so, it defines two main tools, the value function $V$ and the action-value function $Q$:
\begin{align*}
V^\mu(s) &= \mathbb{E_{\mu}}\left[R_t | S_t = s\right] \\
Q^\mu(s, a) &= \mathbb{E_{\mu}}\left[R_t | S_t = s, A_t = a\right]
\end{align*}
\subsection{Deterministic Policy Gradient}
The definition of the $Q$ function allows for a recursive definition as follows:
\begin{equation}
Q^\mu(s, a) = \mathbb{E_{\mu}}\left[r(s,a) + \gamma\mathbb{E}[Q^\mu(s', a')]\right]
\end{equation}
A famous algorithm, Q-learning, assumes that the policy at time $t+1$ is the optimal policy, thus the action selected is the one with the highest expected return, leading to a new definition of the $Q$ function:
\begin{equation}
Q^\mu(s, a) = \mathbb{E}\left[r(s,a) + \gamma \max_a Q^\mu(s', a)\right]
\end{equation}
This reformulation is now dependent only on the environment and thus is off-policy. If we model the greedy policy as a neural network $\mu$ parameterized by $\phi$, the final formulation becomes the following:
\begin{equation}
Q^\mu(s, a) = \mathbb{E}\left[r(s,a) + \gamma Q^\mu(s', \mu_\phi(s'))\right]
\end{equation}
The DPG algorithm \cite{silver2014deterministic_policy_grad} uses this definition to derive an update equation for the policy given a differentiable model $Q_\theta$ via the following equation.
\begin{align*}
    \nabla J(\phi) &\approx \nabla_\phi [Q_\theta(s, a)|_{s = s_t, a = \mu_\phi(s)}] \\
    & = \nabla_a [Q_\theta(s, a)|_{s = s_t, a = \mu_\phi(s)}] \nabla_\phi \mu_\phi(s) |_{s=s_t}
\end{align*}
\subsection{TD3}
TD3 \cite{fujimoto2018td3} is an improvement over DDPG that tries to address mainly the overestimation bias, and introduces three new main components: a double $Q$ estimation, a smoothing of the target for Q-learning, and a slow update.\\
The authors use the double $Q$ estimation in order to address the overestimation bias by changing the target of Q-learning to a conservative one.
\begin{equation}
    y = r + \gamma \text{min}_{i =1,2} Q_{\theta'_i}(s',a'), a' \sim \mu_{\phi'}(s')
    \label{eq:td3_targets}
\end{equation}
where $Q_{\theta'_i}$ and $\mu_{\phi'}$ are two Exponential Moving Averages (EMA) of the actual current neural network, updated at every update step via a linear combination of the EMA and the current solution:
\begin{align*}
\phi'_{new} = \alpha \phi + (1-\alpha) \phi', \alpha \in (0,1] \\
\theta'_{new} = \beta \theta + (1-\beta) \theta', \beta \in (0,1]
\end{align*}

\begin{figure}[t]
    \centering
    \subfloat{\includegraphics[width=\columnwidth]{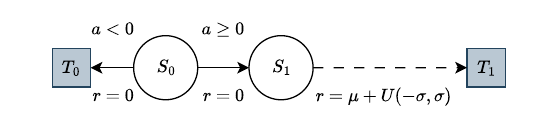}}%
    
    \subfloat{\includegraphics[width=\columnwidth]{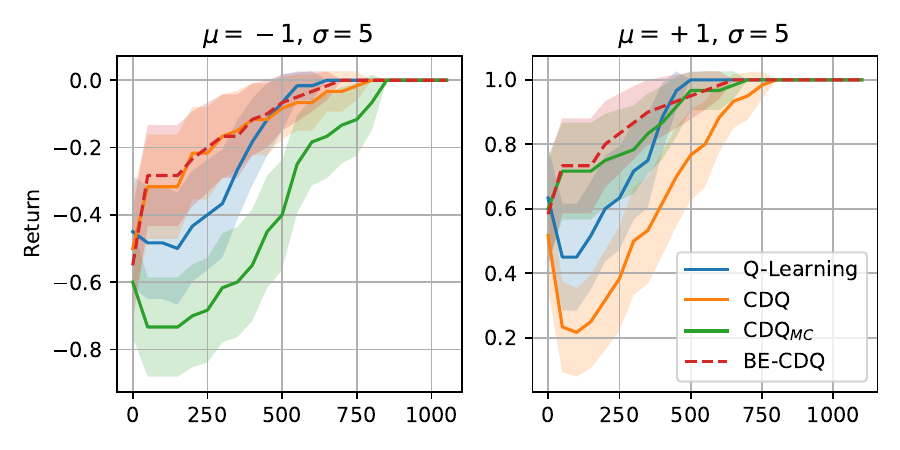}}%
    
    \subfloat{\includegraphics[width=\columnwidth]{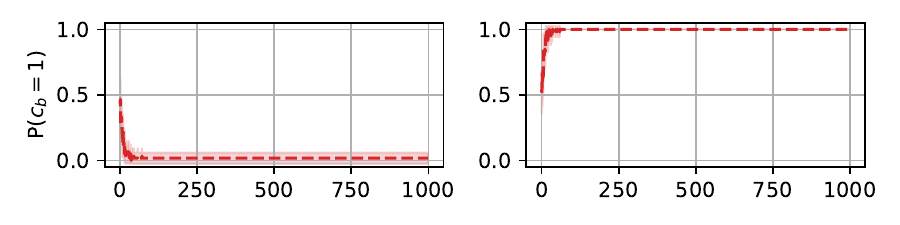}}
    \caption{Above the custom MDP with continuous action space: when $\mu = +1$, $\sigma = 5$ overestimation is favored, with $\mu = -1$, $\sigma = 5$ underestimation is favored. Middle, returns curves of tested algorithms. Below, is the bias choice of the bandit algorithm in BE-CDQ. Plots and shaded areas indicate respectively mean and half a standard deviation from evaluation across 60 random seeds for simulator and network initializations.}
    \label{fig:synth_env}
\end{figure}
\section{Estimation bias in Critic updates with Deep Double Q-Learning}
\label{sec:critic}
In this section, we empirically investigate the effects of estimation bias on the critic updates of CDQ. 
As done in \cite{lan2019maxmin_qlearning} for the case of discrete action space, we want to empirically show that both overestimation and underestimation bias may influence learning performance also in continuous action spaces.
We hypothesize this to be an intrinsic property of the environment and its reward function. 
Q-learning is affected by overestimation bias, since the targets $y = r + \gamma \text{max}_a Q(s', a)$ are used to train the network. However, this bias may not be destructive in all cases. Overestimation bias can help encourage the exploration of highly rewarding actions, whereas underestimation bias might discourage it. Particularly, we refer to the common cases where state-action areas of high reward are highly stochastic or noisy. In these settings, a policy trained with an underestimation bias is less encouraged to explore such areas. At the same time, if areas of low reward are affected by stochasticity or noise, then an overestimation bias might cause the agent to over-explore such places.
These considerations bring us to the investigation presented in this section, we first consider a synthetic environment in which we control the source of noisy rewards that affect the estimation bias of Deep RL algorithms. Secondly, we show that environments of complex dynamics also share similar properties.

\subsection{The effect of estimation bias on Clipped Double Q-Learning}
\label{subsec:synth_env}
To showcase the effect of estimation bias in the critic update phases of CDQ, we use an MDP, represented in the upper part of \cref{fig:synth_env}, heavily inspired by the one proposed in \cite{lan2019maxmin_qlearning}.  Indeed, in \cite{lan2019maxmin_qlearning}, they use this MDP to showcase how, in the discrete action space, overestimation and underestimation can be both suboptimal with small changes to the MDP.\\ 
Our MDP is built as follows. It is composed of four states, $S_0, S_1, T_0, T_1$, with $T_0$ and $T_1$ being terminal states. The agent is initialized in $S_0$. The action space is continuous and defined as $\mathcal{A} = [-1, 1]$. The transition function is deterministic in the whole MDP. When in $S_0$, an action $a < 0$ leads the agent to $T_0$, and an action $a \geq 0$ results in a transition to $S_1$. Once in $S_1$, the agent will move to $T_1$, regardless of the action it selects.
The reward is always $0$ in the whole MDP, except for the transition $S_1 \rightarrow T_1$, where the reward is a random variable $r \sim \mu + U(-\sigma, +\sigma)$. 
This way, $\mathbb{E}[r] = \mu$, therefore if $\mu>0$, then an optimal action is $a > 0$, while if $\mu < 0$ an optimal action is instead $a \geq 0$. However, the introduction of the noise $U(-\sigma, +\sigma)$ makes such reward stochastic, and will eventually lead to different performances of the algorithms. 
Indeed, when $\mu >0$, algorithms that underestimate will be ``\emph{skeptical}'' of such noise, giving more weight to the negative noises than the positive ones, whereas instead, algorithms that overestimate will use such noise to learn faster. The other way around happens when $\mu < 0$. To conclude, we can say, as reported in \cite{lan2019maxmin_qlearning}, a sufficient condition to observe overestimation works better than underestimation is if noisy regions of the MDP correlate with the high reward of such.\\
To see how the estimation bias influences RL algorithms, we simulate two different versions of the environment: (i) an MDP that favors underestimating agents by setting $\mu=-1$ and $\sigma=5$, (ii) an MDP that favors overestimating agents by setting $\mu=+1$ and $\sigma=5$.\\
For each of these MDPs, we test three algorithms derived from Q-Learning and CDQ, each of which trains a Critic network $Q_\theta^S$ for each non-terminal state $S$ of the MDP. The Actor instead is not an additional network trained with a DPG approach but is instead replaced by a simple heuristic that samples $M$ actions $a_1, \dots, a_M$ from $\mathcal{A}$, and then selects $\hat{a} = \text{argmax}_{a_i} \{Q_\theta^S(a_1), \dots, Q_\theta^S(a_M)\}$. This allows us to evaluate the agent's performance as a direct consequence of the training mechanism used for the critic update, without introducing other errors to compute the policy gradient.
Using this approach we train an agent with Q-Learning, therefore the Critic network is a single NN trained with the Q-learning targets, which is known to have an overestimation bias, serving as a baseline agent. We refer to this agent as Q-Learning. A second agent is trained with CDQ, which as stated by the authors \cite{fujimoto2018td3} might introduce underestimation, therefore this agent has two Critic Networks $Q_{\theta_1}^S$ and  $Q_{\theta_2}^S$ for every non-terminal state $S$, each trained with the targets from \cref{eq:td3_targets}, while the estimate for the policy is obtained using $Q_{\theta_1}^S$. This agent is called CDQ.
A third agent is trained using the same networks as the second but with targets computed with the following rule:
\begin{equation}
    y = r + \gamma \text{max}_{i=1,2} Q_{\theta_i}^{S'}(a').
    \label{eq:cdq_max_critic_targets}
\end{equation}
This agent is called CDQ Max Critic (CDQ$_{MC}$).
During training, the actions are applied with an $\varepsilon'$-greedy policy for exploration.
Each agent is tested on both the MDPs running it for a horizon of $10^3$ episodes on 60 different random seeds to minimize as much as possible the effect of the initialization of the networks. The networks are simple MLPs with one $tanh$ hidden layer of $64$ neurons, optimized with Adam \cite{kingma2014adam}, with minibatch size set to $64$. At the end of each episode, the agents' networks are updated with a gradient step. An evaluation is run each 50 episodes, which averages the reward over 10 different episodes, played with $\varepsilon'=0$.\\
We include the evaluations of the trained agents in \cref{fig:synth_env}, where we observe that the behavior pointed out in \cite{lan2019maxmin_qlearning} for the discrete action space setting, is also valid for the continuous action space settings. Indeed, in the MDP with $\mu = +1$, we can observe how the CDQ mechanism struggles to learn compared to the counterparts, where instead, in the case of $\mu = -1$, the $\text{CDQ}_{MC}$ mechanism is the one that struggles the most. We can further observe how indeed, Q-learning falls somewhere in the middle since it's probably overestimating the returns, but in a less aggressive way than $\text{CDQ}_{MC}$, which is explicitly designed to overestimate. 
This experiment reinforces the hypothesis in this paper, that the two estimation biases can hurt or help learning a RL policy, depending on the combination of dynamics and reward function of the environment.
Finally, the method proposed in this paper, based on a dual-armed bandit approach, is applied to the online bias choice between the target computation rule of CDQ and CDQ$_{MC}$. 
In \cref{fig:synth_env} we report its evaluations under the name of BE-CDQ, the algorithm is presented in the next section.
BE-CDQ picks up the optimal bias in the very first iterations and proceeds to have comparable performances to the best of the alternatives in both MDPs, showing its effectiveness in the selection of the correct bias.

\subsection{Estimation bias in complex dynamics}
The property of environments to favor a certain estimation bias is also present in environments of complex dynamics. To highlight this, we consider a set of continuous control environments from OpenAI Gym \cite{OpenAI_gym}.
We present an ablation study obtained by changing the computation of targets for Clipped Double Q-learning in the TD3 algorithm, as done with the custom MDP.
Namely, in TD3 the targets are computed as in \cref{eq:td3_targets}. Here we consider 3 alternative strategies:
\\
TD3 Max Critic (TD3$_{MC}$):
\begin{equation}
    y = r + \gamma \text{max}_{i=1,2} Q_{\theta_i'}(s', a'),
    \label{eq:td3_max_critic_targets}
\end{equation}
with this update, we compute the Q-learning targets using the maximum value between the $Q$ Networks. This increases the overestimation bias and adds no underestimation. This is the exact opposite of TD3's target update.
\\
TD3 Average Critic (TD3$_{AC}$):
\begin{equation}
    y = r + \gamma \frac{1}{2} \left( Q_{\theta_1'}(s', a') + Q_{\theta_2'}(s', a') \right),
\end{equation}

with this update, we compute the Q-learning targets using the mean value between the Q-networks. 
This update doesn't introduce additional estimation bias, and the overestimation bias of Q-learning is possibly reduced.
TD3 Random Critic (TD3$_{RC}$):
        \begin{equation}
        y = r + \gamma \left( \beta Q_{\theta_1}(s', a') + (1-\beta) Q_{\theta_2}(s', a') \right)
    \end{equation}
where $\beta \in {0, 1}$ is a binary mask, sampled uniformly randomly before each episode to select which network is used to compute the targets. Each of these target update strategies is applied to TD3, while the rest of the algorithm is the same as described in \cite{fujimoto2018td3}. 
\begin{figure}[t]
    \centering
    \subfloat{\includegraphics[width=\columnwidth]{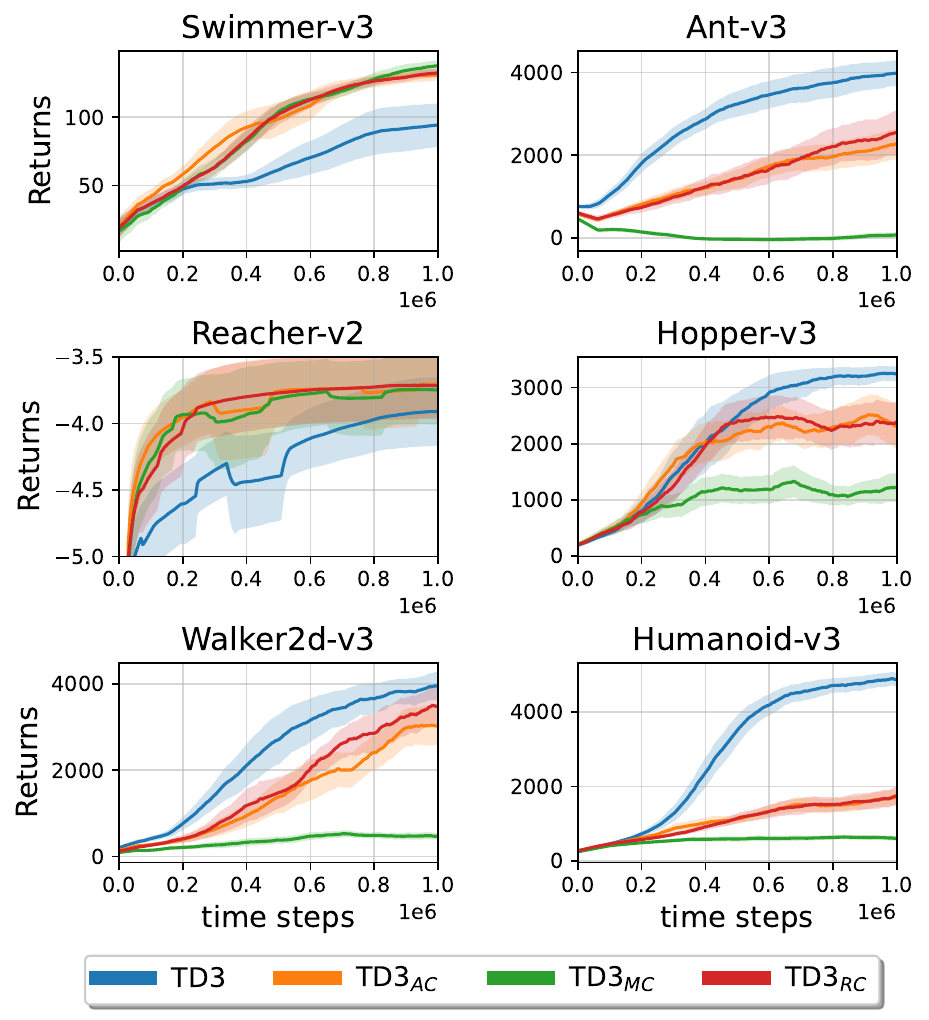}}
    \caption{Training progress curves for continuous control tasks in OpenAI Gym, showing the effect of the different target computations in TD3. Plots and shaded areas indicate respectively mean and half a standard deviation from evaluation across 10 random seeds for simulator and network initializations.}
    \label{fig:ablation_critic}
\end{figure}
This results in creating a different version of TD3 for each strategy.
To evaluate the performance of each version of TD3, we benchmark on the environments from OpenAI Gym.  We report the results of a significant subset of the environments in \cref{fig:ablation_critic}. 
\begin{algorithm}[t]
   \caption{Bias Exploiting TD3}
   \label{alg:be_td3}
\begin{algorithmic}[5]
   \STATE Initialize critic $Q_{\theta_1}, Q_{\theta_2}$, and actor $\mu_\phi$ networks
   \STATE Initialize target networks $\theta_1' \leftarrow \theta_1, \theta_2' \leftarrow \theta_2, \phi' \leftarrow \phi$
   \STATE Initialize replay memory \textbf{$\mathcal{B}$}
   \STATE Initialize bandit problem: $\varepsilon_d$, $\alpha$, $e_r$ 
   \STATE $t=0$, $\varepsilon=0.9$, $Q_b = [0,0]$
   \REPEAT
       \STATE $k=0$, $R=0$, $e=0$, $s \sim p(s_0)$
       \STATE $c_b$ $\leftarrow$ $\varepsilon$-greedy choice from $Q_b$
       \REPEAT
            
            \STATE Select action with exploration noise $a \sim \mu_{\phi'}(s) + \omega$, $\omega \sim \mathcal{N}(0, \sigma)$ and observe $r$ and  $s'$.
            \STATE $done=s'\text{ is terminal}, R = R + r$
            \STATE Store $(s,a,r,s')$ tuple in $\mathcal{B}$
            \STATE Sample mini-batch of $N$ tuples $(s,a,r,s')$ from $\mathcal{B}$
            \STATE $\Tilde{a} \leftarrow \mu_{\phi'}(s') + \omega$,  $\omega \sim \text{clip}(\mathcal{N}(0, \Tilde{\sigma}), -c, c)$
            \IF{$c_b == 0$} 
                \STATE $y = r + \gamma \text{min}_{i=1,2} Q_{\theta_i'}(s', \Tilde{a})$ [\cref{eq:td3_targets}]
            \ELSIF{$c_b == 1$}
                \STATE $y = r + \gamma \text{max}_{i=1,2} Q_{\theta_i'}(s', \Tilde{a})$ [\cref{eq:td3_max_critic_targets}]
            \ENDIF
            
            \STATE Update critics $\theta_i \leftarrow \text{argmin}_{\theta_i} \frac{1}{N} \Sigma (y-Q_{\theta_i}(s,a))^2$
            \IF{$t$ mod $d$}
                \STATE Update $\phi$ by deterministic policy gradient:
                \STATE $\nabla_\phi J(\phi) = \frac{1}{N} \Sigma \nabla_a Q_{\theta_1}(s,a) |_{a=\mu_\phi(s)} \nabla_\phi \mu_{\phi}(s)$
                \STATE Update target networks:
                \STATE $\theta_i' \leftarrow \tau \theta_i + (1-\tau) \theta_i'$
                \STATE $\phi' \leftarrow \tau \phi + (1-\tau) \phi'$
            \ENDIF
            \STATE $k=k+1$
       \UNTIL{not $done$ and $k<K$ and $t+k < T$}
       \STATE $Q_{b|c_b} = Q_{b|c_b} + \alpha (R - Q_{b|c_b})$
       \STATE $t = t+k$, $e=e + 1$
       \STATE \textbf{if} $e \text{ mod } e_r$ \textbf{then} 
       $\varepsilon = 0.9$
       \STATE \textbf{else} $\varepsilon = \varepsilon \cdot \varepsilon_d$
       
   \UNTIL{$t < T$}
\end{algorithmic}
\end{algorithm}
It is possible to observe that the tested algorithms share similar performance in the first steps while becoming noticeably different in the transitory steps, with very distinct final values. We can observe that in the environments \textit{Ant}, \textit{Hopper}, \textit{Walker}, and \textit{Humanoid}, TD3 reaches higher Returns than the other modifications within the available time. In these environments in particular, it is possible to notice a distinct trend: TD3 performs the best, TD3$_{MC}$ performs the worst, while TD3$_{AC}$ and TD3$_{RC}$ both lie somewhere in the middle, with no noticeable difference between the two.\\
Instead, in \textit{Reacher} and \textit{Swimmer}, the relationship is somewhat inverted, with TD3$_{MC}$ performing the best and TD3 performing the worst.
These results indicate that in some environments, the additional overestimation in TD3$_{MC}$ is beneficial for exploring high-rewarding actions. This could be due to environment dynamics, type of reward function, or a combination of both, as shown in the simple MDP in \cref{subsec:synth_env}.
The main result of this ablation study is that none of the considered update strategies performs the best among all the considered environments, including CDQ from TD3. 

\section{Exploiting the Estimation Bias}
\label{sec:bias_exploitation}
Given the considerations and the results presented in the previous section, we look into the idea of learning to exploit the \textit{correct} estimation bias online. With \textit{correct}, we refer to the bias that statistically brings the policy to higher rewards, which doesn't necessarily mean that the value function is closer to the true $Q$.
We propose Bias Exploiting - Twin Delayed Deep Deterministic Policy Gradient (BE-TD3), which extends TD3, adding a decision layer on top of the original algorithm to select which bias to use, namely \cref{eq:td3_targets} or \cref{eq:td3_max_critic_targets}. The algorithm shares the structure of DDPG and TD3, where like TD3, it trains a pair of critics and a single actor. For both the critics and actors, a target copy is maintained and slowly updated to compute the Q-learning targets.
At the beginning of each episode, an $\varepsilon$-greedy policy, guided by a tabular value function $Q_b \in \mathbb{R}^2$, calculates the bias choice $c_b \in \{0,1\}$. In this context, $c_b=0$ denotes underestimation, while $c_b=1$ corresponds to overestimation.
$Q_b$ is initialized to the zero vector at the initial episode, and it is updated using the undiscounted return $R$ obtained at the termination of the episode, with a fixed step size. The scalar $\varepsilon$ is initialized to 0.9, and it decays exponentially each time a decision is taken. 
During an episode, at each time step, we update the pair of critics with the update rule from \cref{eq:td3_targets} if $c_b=0$, else with \cref{eq:td3_max_critic_targets} if instead $c_b=1$. In both cases, the targets are computed using target policy smoothing. 
The actor is updated every $d$ steps (delayed updates), with respect to $Q_{\theta_1}$ following the deterministic policy gradient algorithm \cite{silver2014deterministic_policy_grad}.
We test the bandit approach first to the bias selection problem in the simple MDP presented in \cref{subsec:synth_env}, we apply the BE approach to the CDQ approach described in the experiment of \cref{fig:synth_env}. The bandit is used to choose which rule to use to compute the Q-learning targets for the two networks, we refer to this approach as Bias Exploiting - Clipped Double Q-learning (BE-CDQ). 
Its results are included in \cref{fig:synth_env}, reporting its return curve and the plot of the action chosen by the bandit algorithm. The choice curves show that in the MDP that favors underestimation ($\mu=-1$, $\sigma=5$), BE-CDQ chooses $c_b=0$ after the first episodes, consequently, its return curve follows CDQ. Instead, in the MDP that favors overestimation ($\mu=+1$, $\sigma=5$) BE-CDQ chooses $c_b=1$ after the first episodes, and the return curve follows CDQ$_{MC}$. 

\begin{figure}[h]
    \centering
    \includegraphics[width=\columnwidth]{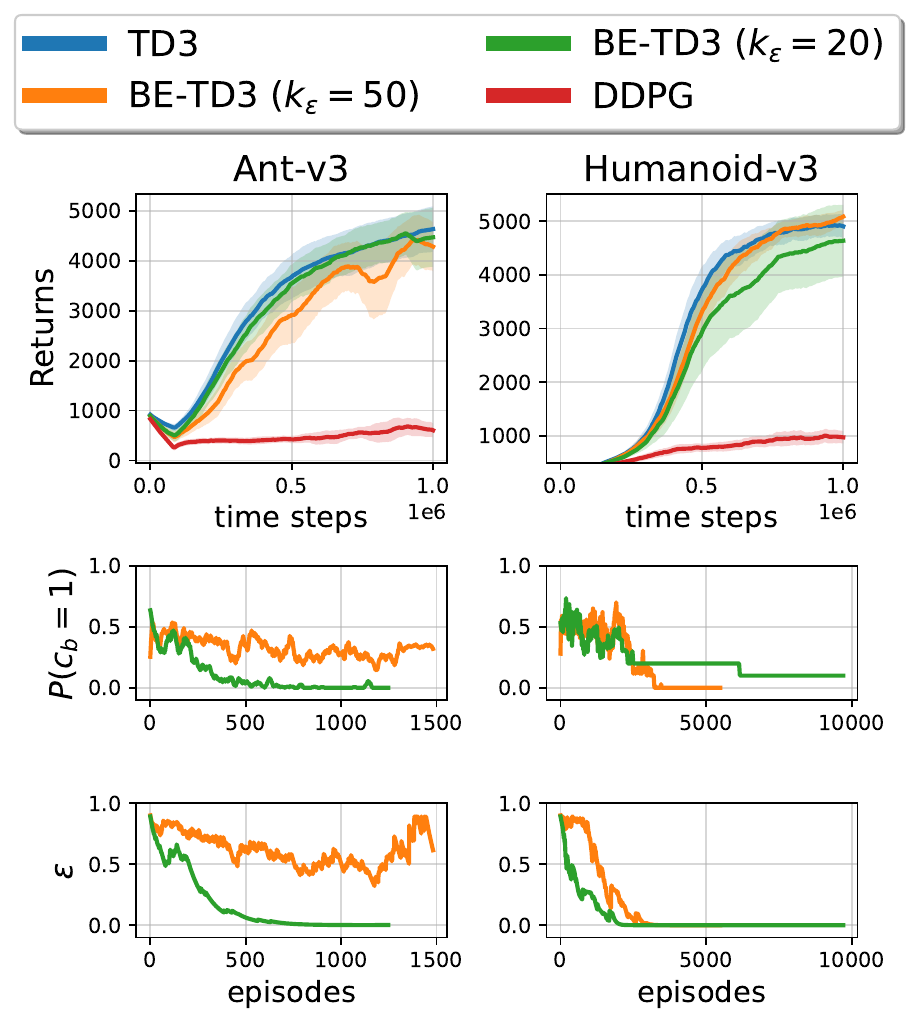}
    \caption{Training progress curves showing the effect of the soft $\varepsilon$ reset scheduling in BE-TD3. Plots are from 10 random seeds for simulator and network initializations, smoothed for visualization. Evaluations of Return are performed every 5000 time steps, plots show mean and half a standard deviation, over the seeds. ($\varepsilon_d=0.99$, $\alpha=0.25$)}
    \label{fig:soft_eps_reset_example}
\end{figure}

\subsection{Non-stationary bandit problem}
The bandit in BE-TD3 is subject to strongly changing dynamics, due to the updates of the actor, resulting in strong non-stationarity. This translates to the following consideration: if a bias is preferred during the initial episodes, it might not be the best choice during the latest interaction time.
To deal with this, we reset $\varepsilon$ to its initial value for each $k_\varepsilon$ episodes. Namely, once every $k_\varepsilon$ episodes, the algorithm explores again the bias choice. The $\varepsilon$ decay scheduling is an important aspect of the algorithm, as it can strongly influence performance. The following update rule describes an alternative scheduling of the $\varepsilon$ decay
\begin{equation}
    \varepsilon = \text{min}(\varepsilon \cdot \varepsilon_d \cdot \text{max}(\frac{k_\varepsilon}{k}, 1), 0.9),
\end{equation}
which substitutes lines 31 and 32 of \cref{alg:be_td3}. This rule performs a soft reset of $\varepsilon$ instead of a hard one, delaying the decay until episodes are shorter than $k_\varepsilon$ steps.
This helps in environments with complex dynamics, most notably in \textit{Humanoid}, but in turn, $k_\varepsilon$ must be sensibly tuned.
In \cref{fig:soft_eps_reset_example} we show the benchmarks of BE-TD3 with this alternative $\varepsilon$ scheduling for $k_\varepsilon=50$ and $k_\varepsilon=20$. With $k_\varepsilon=20$ BE-TD3 works particularly well in \textit{Humanoid}, but deteriorates performances in \textit{Ant} with respect to TD3. With $k_\varepsilon=20$ we fix the instability issues in \textit{Ant}, but slow down learning in \textit{Humanoid}. Even if the soft $\varepsilon$ update rule seems intriguing, in practice results are too sensible to the hyperparameters $k_\varepsilon$, which should be tuned separately to each environment. Therefore, the hard reset strategy is preferred in this paper.

\subsection{Exploration of estimation bias}
\begin{figure}[h]
    \centering
    \includegraphics[width=\columnwidth]{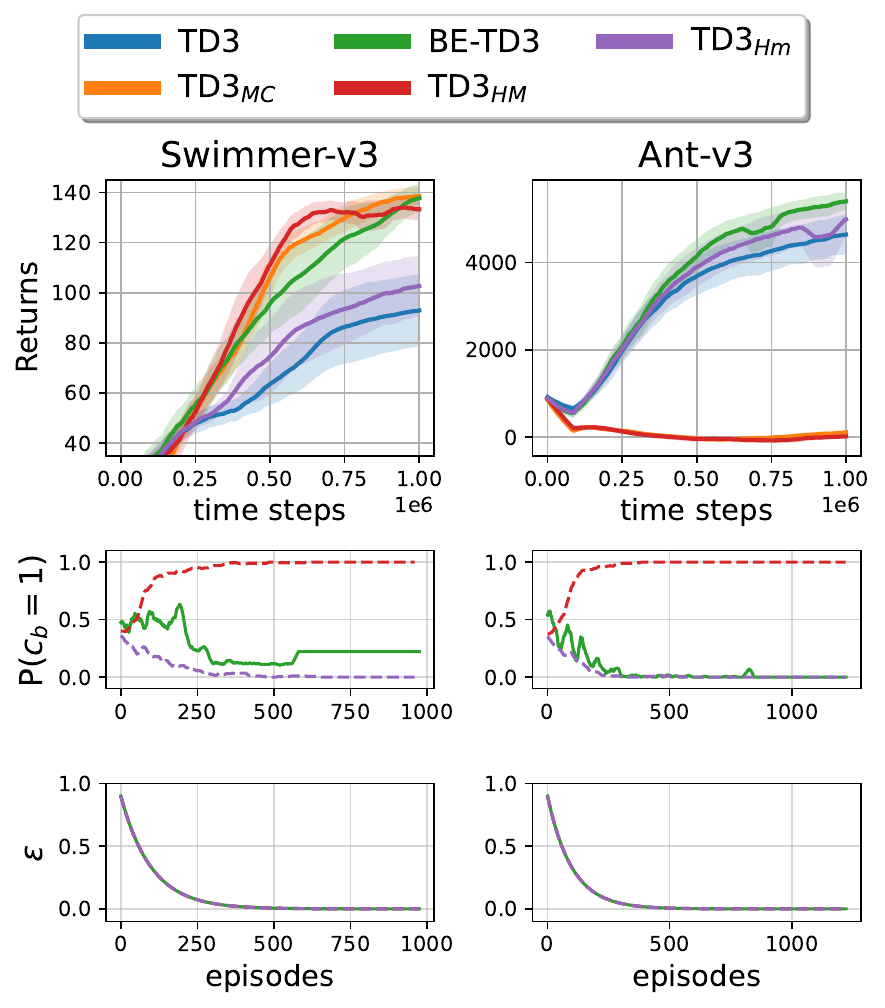}
    \caption{Comparing bias exploitation in \textit{Swimmer}, \textit{Ant}, and \textit{Hopper}. 
    Plots are from 10 random seeds for simulator and network initializations, smoothed for visualization. Evaluations of Return are performed every 5000 time steps, plots show mean and half a standard deviation, over the seeds. In these experiments, $\varepsilon$ is not reset ($\alpha=0.25$, $\varepsilon_d=0.99$). }
    \label{fig:comparing_epsilon_greedy_to_heuristic_swimmer_ant_hopper}
\end{figure}

The $\varepsilon$-greedy choice of estimation bias in the algorithm BE-TD3 could be misinterpreted as just a technique of injection of random noise for exploration. 
In \cref{fig:comparing_epsilon_greedy_to_heuristic_swimmer_ant_hopper} we compare BE-TD3 to TD3, TD3$_{MC}$ and two versions of BE-TD3, where the estimation bias is chosen between \cref{eq:td3_targets} and \cref{eq:td3_max_critic_targets} following an heuristic rule instead of the bandit strategy. Namely:\\
\begin{itemize}[topsep=0pt,parsep=0pt,partopsep=0pt]
    \item TD3 Heuristic Max (TD3$_{HM}$): like BE-TD3, follows a $\varepsilon$-greedy policy to select between underestimation and overestimation, but the greedy choice is always overestimation (\cref{eq:td3_max_critic_targets}). This baseline is TD3$_{MC}$ with some exploration of estimation bias in the first episodes;
    \item TD3 Heuristic min (TD3$_{Hm}$): like BE-TD3, follows a $\varepsilon$-greedy policy to select between underestimation and overestimation, but the greedy choice is always underestimation (\cref{eq:td3_targets}). This baseline is TD3 with some exploration of estimation bias in the first episodes: 
\end{itemize}
In \cref{fig:comparing_epsilon_greedy_to_heuristic_swimmer_ant_hopper}, we compare TD3, TD3$_{MC}$, BE-TD3, TD3$_{HM}$ and TD3$_{Hm}$ on \textit{Swimmer} and \textit{Ant}. TD3$_{Hm}$ only slightly improves the final average performance of TD3 in Swimer, an environment that benefits from overestimation. Instead, it improves in \textit{Ant}. This experiment seems to confirm the hypothesis that an initial exploration of the estimation bias could bootstrap policy learning in these environments. At the same time, it shows that still, the bandit algorithm selecting a meaningful bias is required to reach the best performances.

We report the pseudocode of BE-TD3 in \Cref{alg:be_td3} and its benchmarks in \cref{fig:benchmarks_BETD3}. Results are commented in the end of the next section to provide a more exhaustive outlook.
The very low computational burden added by the bandit approach makes it very easy and convenient to apply it to most Actor-Critic approaches.
We can also apply the bias exploitation to SAC, which shares a lot of structure with TD3. We refer to this new version of SAC as Bias Exploiting - Soft Actor Critic (BE-SAC), we omit to include its pseudocode, as it is almost identical to \cref{alg:be_td3}. 
Finally, treating the bias selection problem as a Bandit can seem naive, as more sophisticated approaches exist. For example, the same problem could be treated as a classification problem.
Nonetheless, since bias selection is a stateless classification problem, using these simple alternatives is preferred to minimize the possible influence of other factors on policy optimization.

\subsection{Benchmarks and Results}
\begin{figure*}[h!]
    \centering
    \subfloat{\includegraphics[width=0.98\textwidth]{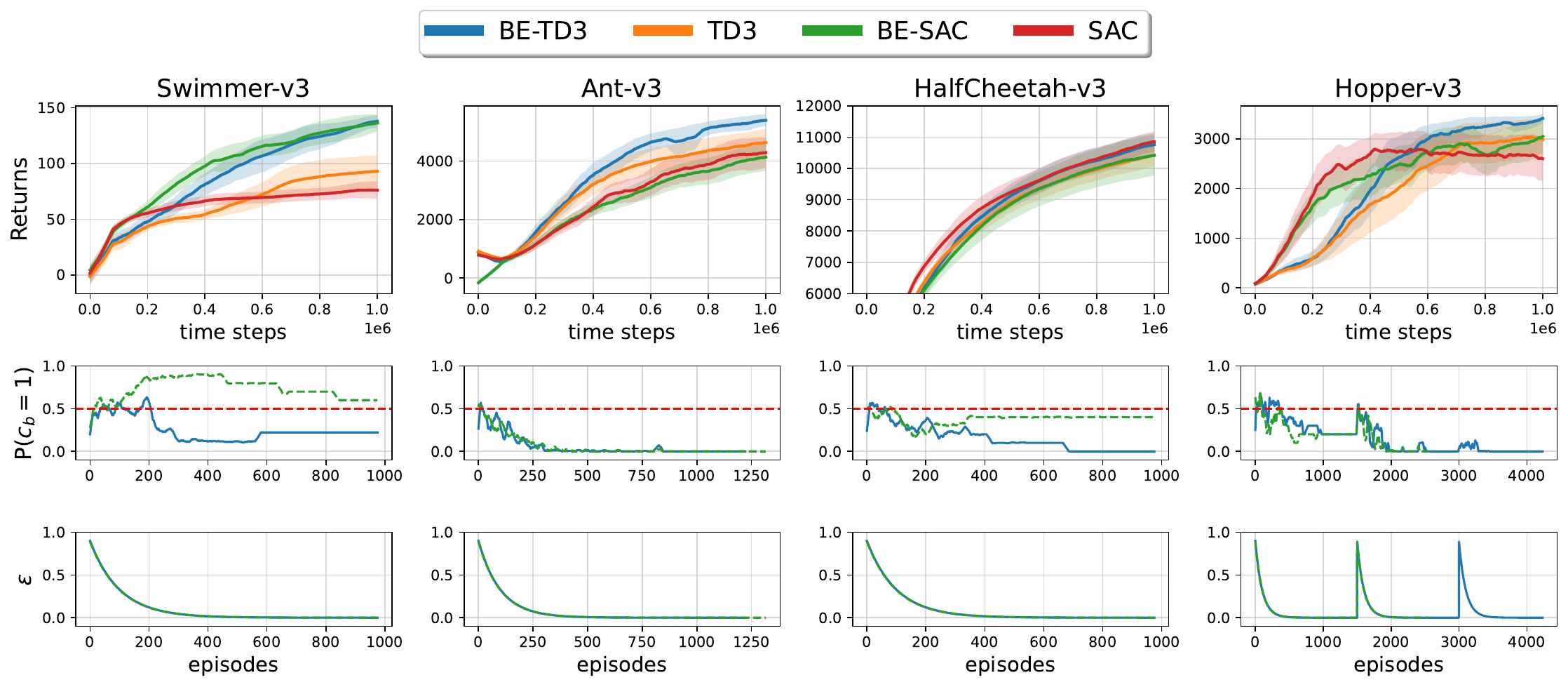}}
    
    \subfloat{\includegraphics[width=\textwidth]{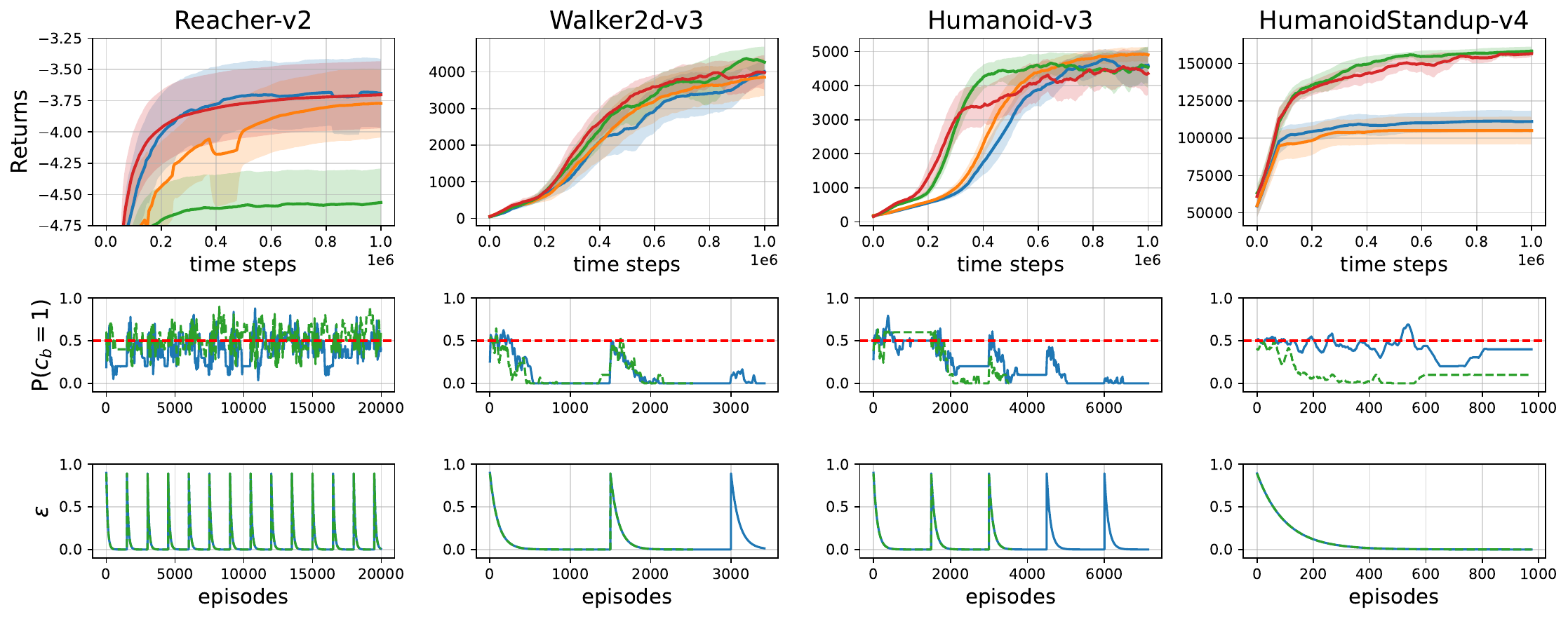}}
    \caption{
    Comparing Bias Exploiting TD3 with baselines in continuous control tasks. 
    Plots are from 10 random seeds for simulator and network initializations, smoothed for visualization. Evaluations of Return are performed every 5000 time steps, plots show mean and half a standard deviation of evaluation over 10 episodes. For each environment, we report the probability of the bandits choosing overestimation and $\varepsilon$. (In all BE algorithms $\varepsilon_d=0.99$, $e_r=1500$, $\alpha=0.25$)}
    \label{fig:benchmarks_BETD3}
\end{figure*}

Finally, we evaluate the BE algorithms in a set of continuous control tasks from OpenAI. For each environment, we run 10 instances of BE-TD3, BE-SAC, TD3, and SAC for different random seeds, in order to have reliable statistics.
In each instance, we run the algorithm for a maximum of $10^6$ steps, evaluating the policy for each $5 \cdot 10^3$ step. The evaluations are performed by testing the current policy on the environment, without exploration noise, 10 times, and collecting the average cumulative return. 
For the BE algorithms, we also save the bandit choice and $\varepsilon$ throughout the episodes.
This gives us the results reported in \cref{fig:benchmarks_BETD3}, for each task we report the statistics of the average returns, as well as the mean of the bandit choices, which represent the probability of the agent of choosing overestimation during the training process. Furthermore, the plots include the $\varepsilon$ scheduling.
We also compute the maximum evaluations among the various seeds for each task and algorithm and report the statistics in \cref{table:max_avg_returns}, where the highest rewards are highlighted.
The return curves show that the BE algorithms surpass their baseline counterpart, with the only exceptions of BE-SAC in \textit{Reacher} and BE-TD3 in \textit{Humanoid}. Moreover, the plots of the bandit choice show that, generally, the bias choices go toward the expected option. In particular, in the final steps of training in \textit{Ant}, \textit{Hopper}, \textit{Walker}, and \textit{Humanoid} all the agents choose underestimation, while in the other tasks, the agents select overestimation with a certain probability. The statistics in \cref{table:max_avg_returns} confirm that in almost all cases, the bandit approach in BE-TD3 improves the final policy performance. 

\begin{table*}[h]
\caption{Maximum Average Return over 10 evaluations, across 10 trials, each consisting of 1 million time steps. The bolded values represent the peak performance for each task.}
\label{table:max_avg_returns}
\vskip 0.15in
\begin{center}
\begin{small}
\begin{sc}
\begin{tabular}{lcccccccr}
\toprule
\textbf{Environment} & \textbf{BE-TD3} & \textbf{TD3} & \textbf{BE-SAC}  & \textbf{SAC}  \\
\midrule
\textit{Swimmer} & \textbf{145.20 $\pm$ 13.25} & 97.18 $\pm$ 29.70 & \textbf{144.64 $\pm$ 15.83} & 83.28 $\pm$ 18.29\\
\textit{Ant} & \textbf{5648.61 $\pm$ 269.48} & 4872.64 $\pm$ 815.51 & 4551.23 $\pm$ 770.67 & 4826.06 $\pm$ 855.06\\
\textit{HalfCheetah} & \textbf{10898.82 $\pm$ 704.67} & 10562.71 $\pm$ 743.48 & 10625.37 $\pm$ 1372.06 & 11051.23 $\pm$ 652.29\\
\textit{Hopper} & \textbf{3503.52 $\pm$ 144.06} & 3230.12 $\pm$ 499.67 & 3484.06 $\pm$ 118.21 & 3421.23 $\pm$ 103.93\\
\textit{Reacher} & \textbf{-3.61 $\pm$ 0.54} & -3.72 $\pm$ 0.56 & -4.45 $\pm$ 0.54 & -3.69 $\pm$ 0.54\\
\textit{Walker2d} & 4230.83 $\pm$ 562.94 & 4089.16 $\pm$ 977.63 & \textbf{4665.94 $\pm$ 394.46} & 4390.64 $\pm$ 607.53\\
\textit{Humanoid} & 5346.18 $\pm$ 105.93 & 5349.46 $\pm$ 129.80 & 5326.68 $\pm$ 79.85 & \textbf{5423.65 $\pm$ 61.61}\\
\textit{HumanoidStandup} & 112441.75 $\pm$ 15667.96 & 105913.19 $\pm$ 18656.28 & \textbf{161581.75 $\pm$ 2801.7}8 & 159280.99 $\pm$ 1865.63\\
\bottomrule
\end{tabular}
\end{sc}
\end{small}
\end{center}
\vskip -0.1in
\end{table*}

\section{Conclusions and future work}
\label{sec:conclusions}
In conclusion, this paper presents significant advancements in the field of Reinforcement Learning (RL) by exploring and exploiting estimation biases in Actor-Critic methods using Deep Double Q-Learning.  

We demonstrate that Online Bias Exploitation in Clipped Double Q-learning can enhance TD3's and SAC's performance, particularly in scenarios where action value overestimation is advantageous. The proposed improvements impose minimal to no computational overhead and could be integrated into various actor-critic methods.
This suggests the potential for opening new possibilities for more computationally efficient algorithms while maintaining high performance in various RL environments

Potential future research avenues involve investigating bias selection in Deep RL algorithms that make use of ensembles of value functions. Within these scenarios, addressing the bias selection problem could be approached as a continuous control task.

\section{Broader Impact}
\label{sec:impact}
This paper presents work whose goal is to advance the field of Deep Reinforcement Learning. There are many potential societal consequences of our work, none of which we feel should be specifically highlighted here.

\bibliography{ref}
\bibliographystyle{icml2024}

\newpage
\appendix
\onecolumn

\section{Networks architectures and hyper-parameters}
For TD3, DDPG, and BE-TD3, the following network architecture have been used for the actor:  \\
\texttt{
(state dim -> 256)\\
Relu\\
(256 -> 256)\\
Relu\\
(256 -> action dim)\\ 
tanh
}
\newline
For SAC and BE-SAC, the following network architecture has been used for the actor:\\
\texttt{
(state dim -> 256)\\
Relu\\
(256 -> 256)\\
Relu\\
(256 -> action dim * 2)
}
\newline\newline
For all methods, the following network architecture has been used for the critic: \\
\texttt{
(state dim + action dim-> 256)\\
Relu\\
(256 -> 256)\\
Relu\\
(256 -> 1)
}

\begin{table*}[h]
\caption{List of hyperparameters used for training.}
\label{table
}
\vskip 0.15in
\begin{center}
\begin{small}
\begin{sc}
\begin{tabular}{lcccccc}
\toprule
\textbf{Hyper-parameter} & \textbf{TD3} & \textbf{SAC} & \textbf{DDPG} & \textbf{BE-SAC} & \textbf{BE-TD3} \\
Critic learning-rate & $0.0003$ & $0.0003$ & $0.0003$ & $0.0003$ & $0.0003$ \\
Actor learning-rate& $0.0003$ & $0.0003$ & $0.0003$ & $0.0003$ & $0.0003$ \\
Optimizer& Adam & Adam & Adam & Adam & Adam \\
Target Update Rate ($\tau$)& $0.005$ & $0.005$ & $0.005$ & $0.005$ & $0.005$ \\
Batch-size& $256$ & $256$ & $256$ & $256$ & $256$ \\
Discount factor & $0.99$ & $0.99$ & $0.99$ & $0.99$ & $0.99$ \\
Exploration policy & $N(0, 0.2)$ & learnt & $N(0, 0.2)$ & learnt & $N(0, 0.2)$ \\
Entropy & - & 0.5 & - & 0.5 & - \\
Actor update delay ($d$) $\dagger$ & $2$ & $1$ & $1$ & $1$ & $2$ \\

\bottomrule
$\dagger$ $d=1$ means no delay
\end{tabular}
\end{sc}
\end{small}
\end{center}
\end{table*}



\end{document}